\documentclass[conference]{IEEEtran}
\usepackage{array}
\usepackage{booktabs}
\usepackage{textcomp}
\usepackage[pdftex]{graphicx}
\graphicspath{{../latex/images/}}
\DeclareGraphicsExtensions{.png}

\usepackage{multirow}
\usepackage{enumerate}
\usepackage{graphicx}
\usepackage[sort&compress,numbers]{natbib}
\usepackage{calrsfs}

\usepackage{amsmath,amssymb}
\usepackage{mathtools}
\usepackage{lmodern}
\usepackage{enumerate}
\usepackage{geometry}
\usepackage{bm}
\usepackage{xcolor,lipsum,subcaption}
\DeclareMathAlphabet{\pazocal}{OMS}{zplm}{m}{n}
\newcommand{\Lb}{\pazocal{L}}

\DeclareMathAlphabet{\mathpzc}{OT1}{pzc}{m}{it}
\makeatletter
\newcommand{\ostar}{\mathbin{\mathpalette\make@circled\star}}
\newcommand{\make@circled}[2]{%
\ooalign{$\m@th#1\smallbigcirc{#1}$\cr\hidewidth$\m@th#1#2$\hidewidth\cr}%
}
\newcommand{\smallbigcirc}[1]{%
\vcenter{\hbox{\scalebox{0.77778}{$\m@th#1\bigcirc$}}}%
}

\newcolumntype{M}[1]{>{\centering\arraybackslash}m{#1}}

\usepackage{multirow}
\usepackage{url}
\ifCLASSINFOpdf
\else
\fi

\begin{document}
%
\title{OccRobNet : Occlusion Robust Network for Accurate 3D Interacting Hand-Object Pose Estimation}
\author{\IEEEauthorblockN{Mallika Garg}
	\IEEEauthorblockA{\textit{ECE Department} \\
		\textit{IIT Roorkee, India }\\
		mallika@ec.iitr.ac.in}
		\and
	\IEEEauthorblockN{Pyari Mohan Pradhan}
	\IEEEauthorblockA{\textit{ECE Department} \\
		\textit{IIT Roorkee, India }\\
		pmpradhan@ece.iitr.ac.in}
	\and
	\IEEEauthorblockN{Debashis Ghosh}
	\IEEEauthorblockA{\textit{ECE Department} \\
		\textit{IIT Roorkee, India }\\
		debashis.ghosh@ece.iitr.ac.in}

}


%


\maketitle

\begin{abstract}
Occlusion is one of the challenging issues when estimating 3D hand pose. This problem becomes more prominent when hand interacts with an object or two hands are involved.  In the past  works, much attention has not been given to these occluded regions. But these regions contain important and beneficial information that is vital for 3D hand pose estimation.  Thus, in this paper, we propose an occlusion robust and accurate method for the estimation of 3D hand-object pose from the input RGB image.  Our method includes first localising the hand joints using a CNN based model and then refining them by extracting contextual information. The self attention transformer then identifies the specific joints along with the hand identity. This helps the model to identify the hand belongingness of a particular joint which helps to detect the joint even in the occluded region. Further, these joints with hand identity are then used to estimate the pose using cross attention mechanism. Thus, by identifying the joints in the occluded region, the obtained network becomes robust to occlusion. Hence, this network  achieves state-of-the-art results when evaluated  on the InterHand2.6M, HO3D  and H$_2$O3D datasets. 
\end{abstract}


%
\IEEEpeerreviewmaketitle

\section{Introduction}\label{sec:intro}
The objective of the hand pose estimation task is to localize the hand joints which are extracted using different methods~\cite{10636024}. This task can be accomplished using a single  RGB image or depth maps. Recently, many deep learning based-methods have been developed with different Convolutional Neural Network (CNN)-based and  Transformer based architectures.  In~\cite{moon2020i2l}, instead of directly predicting hand pose from input image, images are first mapped to lixels. Then, mesh vertices are estimated from per-lixel likelihood. This helps to develop the spatial relationship between the input image and the predicted values. 

Estimating hand pose is an ambiguous problem  that requires a large  dataset~\cite{zimmermann2019freihand, moon2020interhand2,hampali2022keypoint}. It should include various annotations like hand region segmentation mask, hand pose annotations, etc.  These datasets can be single hand pose datasets, single hand interaction with object and two hand interaction with object datasets.  G. Moon et al.~\cite{moon2020interhand2} proposed multi-view hand dataset for single and interacting hands. This dataset is InterHand2.6M which consists of RGB images which are captured with various poses with 80 to 140 high-resolution cameras. It also includes automatic and manual 3D keypoint annotations along with handedness,  2.5D hand pose, and relative depth.     However, C. Zimmermann et al.~\cite{zimmermann2019freihand} proposed a dataset named Freihand. It is the first large scale, multi-view dataset with 3D hand pose and shape annotations.
\begin{figure*}[ht]\centering
\includegraphics[scale=.5]{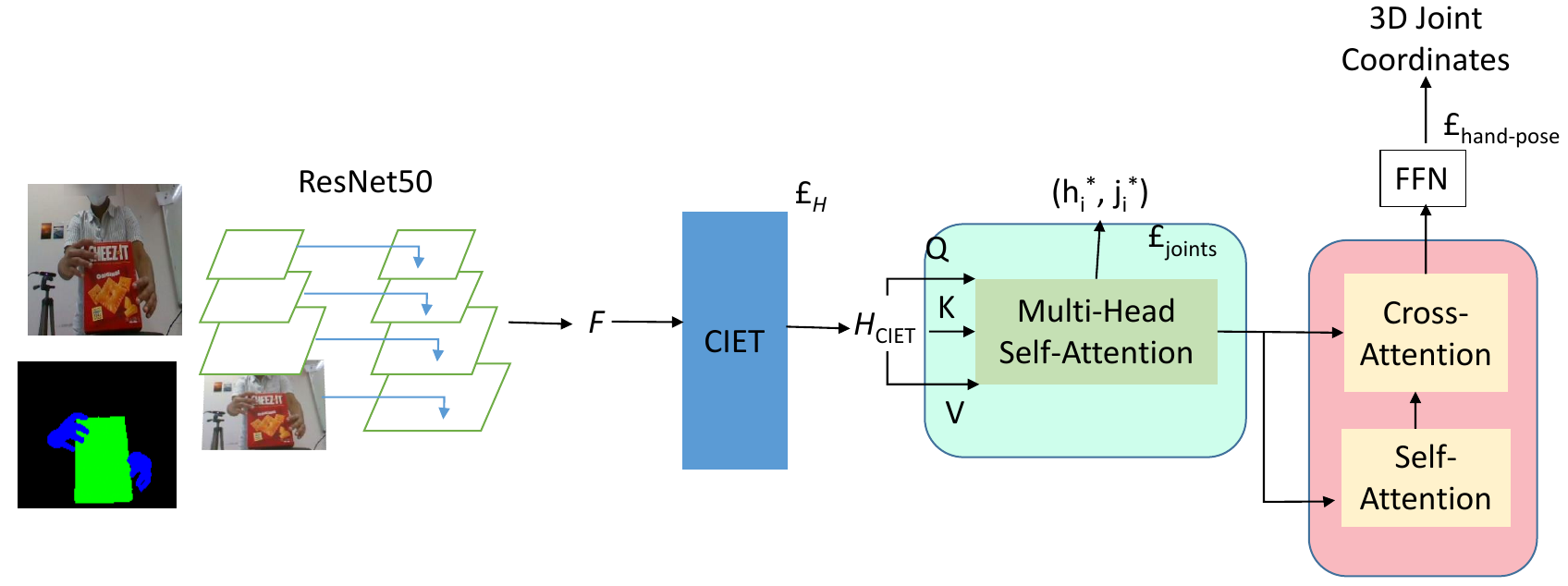}
\caption{Overview of our method. We first estimates 1D heatmaps from RGB hand image and hand-object segmentation mask~\ref{sec:1}, using ResNet50 network.  From the feature maps $F$, the CIET module, which is a transformer module, creates context-aware features which are the refined  object and hand features~\ref{sec:2}. Then,  transformer encoder network is utilized to give attention to hand and object features using sigmoid and softmax attentions. The decoder  cross-attention module finally predicts the hand pose~\ref{sec:3}.} 
\label{fig2}
\end{figure*}  

In this work, we focus on the issue of occlusion which is most common in interacting hands and hand-object interactions. The obtained hand and object features are refined by extracting contextual information using a transformer model. This helps to obtain the hand information even when the hand is occluded. Our architecture is based on Detr model~\cite{zhu2020deformable}. Also, while estimating the hand pose more attention should be given on the hand regions, so we have employed a combination of sigmoid and    softmax attentions in the transformer encoder and decoder network. We have performed various experiments to test the performance of our model on three challenging hand pose datasets including InterHand2.6M~\cite{moon2020interhand2}, HO-3D\_V3~\cite{hampali2020honnotate}  and  H$_2$O-3D~\cite{hampali2022keypoint} dataset and found that our method has state-of-the-art results.

Thus, the contributions of this work are highlighted below:
\begin{enumerate}
\item We propose OccRobNet, an occlusion robust network for estimating 3d hand-object pose by explicitly modeling the relationship between keypoints resulting in more accurate poses from the input RGB image.

\item Our model estimates the hand pose in four stages: extracting context-aware features, detecting  keypoints, associating each keypoint with corresponding joint using self attention mechanism and finally predicting each joint using cross-attention. 

\item We proposed the novel extraction of context-aware feature from the heatmaps obtained using the proposed novel  Contextual Information Enhancement Transformer (CIET) module enabling more reliable 3D hand pose estimation.

\end{enumerate}

\section{Literature survey}
Hand pose estimation,  generally follow model-based techniques utilizing data depth for reconstruction of hand pose. Infrared markers, color-coded gloves, and electrical sensing equipment based techniques are commonly used for accurate and easy pose estimation. The main issues while hand pose annotations with visible regions of hand lies in self-occlusions which is tackled using multiple-view based approaches. Iterative bootstrapping, integrating pose and shape annotations are some of the approaches presented in literature. Further, the most recent is the neural network-based method which  mainly utilizes discriminative methods~\cite{hasson2019learning} to generalize image seen and directly localizes joints. The technique of integrating 2.5D right-left hand poses and relative depth among interacting hand from the input image is utilized for estimation of interacting hand pose from a single RGB image.

S. Baek et al.~\cite{baek2020weakly} presented a generative network based 2D pixel level guidance and mesh renderer based 3D mesh guidance. Input images are segmented and hand-only images for accurate and efficient hand pose estimation.  This method estimates the hand pose without using hand object labels. A similar architecture is presented in~\cite{wei2016convolutional}, but it handles occluded areas in the human body using global scene context. Also, F. Mueller et al.~\cite{mueller2018ganerated} presented a similar method for 3D hand-object interaction. However, D Goudie et al. uses hand segmentation masks from the hand-object interaction images and uses a two stage approach~\cite{goudie20173d}, it doesn't handle occlusion.

Hand pose are sometimes estimated using parametric hand models which is usually MANO~\cite{romero2022embodied},or SMPL- based~\cite{loper2015smpl} model that maps hand pose and shape to hand mesh and generates a hand model. In the literature, there are many methods that estimate 3D hand pose and finally reconstructs the hand model using the predicted hand pose which are mapped to MANO parameters. A self-supervised model for 3D hand reconstruction  can jointly estimate pose, shape, and the camera viewpoint~\cite{chen2021model} from single view RGB image. 2D keypoints are estimated which are then used to estimate 3D joints using 2D-3D consistency loss for the projected joints. Moreover, weak supervision is also used for hand mesh reconstruction~\cite{kulon2020weakly}. It employs an encoder-decoder based system that reconstructs the mesh directly from image coordinates.

\section{Method}\label{sec:formatting}
Our hand pose estimation network first estimates 2D heatmaps, from the given input RGB  hand image. Hand features are extracted from UNet~\cite{ronneberger2015u} network which are then refined by a transformer model that extracting contextual information. These refined heatmaps are then associated with the hand joints using a self attention module which is encoded with their spatial location. The complete flow of our architecture is shown in Fig.~\ref{fig2}.

\subsection{Backbone}\label{sec:1}
The backbone network consists of a ResNet50~\cite{he2016deep} based Feature Pyramid Network (FPN)~\cite{lin2017feature} to extract a pyramid hierarchy of hand feature map, $F\in \mathbb{R}^{H\times W\times C}$. We first extract heatmaps $\mathcal{H}$ from the feature map extracted to localize the  2D hand joints. The ground truth heatmap $\mathcal{H}^*$ is obtained as in~\cite{hampali2022keypoint}. It outputs 2D heatmaps for each joint $j \in \mathcal{J}^{2D}$, where $\mathcal{J}^{2D} \in \mathbb{R}^{N \times 2}$, where $N$ = 21 is the number of hand joints for each hand according to the MANO model. The predicted heatmap of each hand joint has dimension $32 \times 32$. 

Along with the heatmaps, object segmentation is also obtained from the ResNet backbone by simply adding a prediction head to the ResNet model. 20 object points are randomly selected from the   segmentation maps. These object-pose points encode the appearance information of the object for predicting the 3D rotation and translation of the object from the mesh projection  of the 2D object points. 

\begin{figure}[t]\centering
\includegraphics[scale=.42]{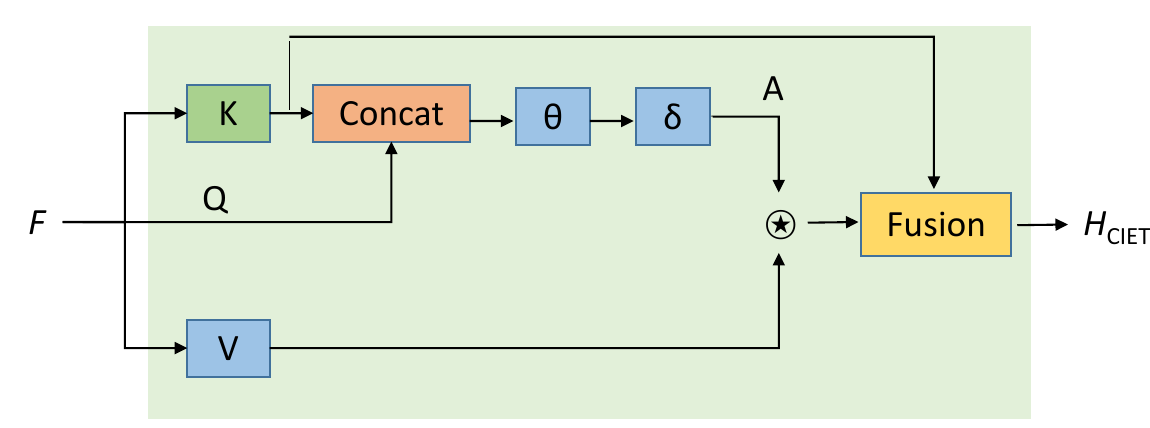}
\caption{The detailed structure of our Contextual Information Enhancement Transformer (CIET) block. $\star$ denotes the  local matrix multiplication and $\theta$ and $\delta$, are used to denote the two 1 $\times$ 1 convolution.}
\label{fig30}
\end{figure}

\subsection{Contextual Information Enhancement Transformer (CIET)}\label{sec:2}
Although, FPN is able to extract multilevel features from the hand image. Still, some rich hand contents are   left under-exploited.  The receptive fields in convolution layers of FPN  hinders the modeling of global features and long term dependencies. So, we have designed a network that extracts the contextual information from the features obtained from the FPN network.    Also,  we have employed an attention mechanism that acts as an enhancement to refine the features obtained from the backbone network. Combining both these needs of attention mechanism and rich contextual information, we have proposed a CIET network that exploits rich contextual information of the feature maps from backbone network using transformer model. 

We have exploited the idea of CoTNet~\cite{li2022contextual} that extract contextual information and integrates it with the self attention learning. Instead of directly feeding the CoTNet with the input RGB image, we have  injected the feature maps $F$  directly into the Query ($Q$), Key ($K$) and Value ($V$).  This can be intuitively thought as looking for some related hand information in entire feature map $F$ as the softmax attention in transformer pays more attention to the relative hand information in the entire image.   This hand information could be the information occluded by the other hand or object, that make the complete system robust to occlusion. The correlation map obtained by CIET is highly correlated to the hand features, since we adopt softmax based attention. Since CIET is an enhancing network, it is backed up by a backbone residual network as mentioned in the previous section. The detailed structure of the CIET network is shown in Fig.~\ref{fig30}.
\begin{figure}[t]\centering
\includegraphics[scale=.48]{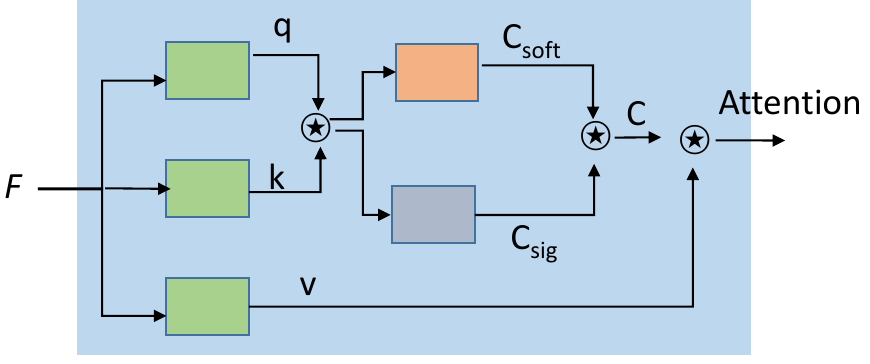}
\caption{The overall structure that combines the softmax attention with the sigmoid attention for the same input query- key pairs. }
\label{fig3}
\end{figure}
Conventional transformer~\cite{vaswani2017attention} employs pairwise query-key scaled-dot product for calculating attention. In this work, instead of directly performing scaled-dot product on query-key pair, we first contextualize the keys by performing $3 \time 3$ convolution over all the neighbouring keys.  The contextualized keys $K^1 \in \mathbb{R}^{H\times W \times C}$ are the  static context among the local neighbours of the input feature $F$. These contextualized keys ($K^1$) are concatenated with the query $Q$, which is then followed by two consecutive $1 \times 1$ convolution. The final attention obtained is represented as
\begin{equation}
\label{eqn:86}
\begin{aligned}
A = [K^1 , Q]W_\theta W_\delta,
\end{aligned}
\end{equation}
where $\theta$ and $\delta$, are used to denote the two $1 \times 1 $ convolution, one with ReLU activation function and the other without activation function.   

Since attention $A$ is calculated from the contextualized key, this attention is called contextualized attention, which is then used to aggregate all the values (V). The final output is the dynamic contextual information of the input feature map which is represented as
\begin{equation}
\label{eqn:88}
\begin{aligned}
\mathcal{H}_{CIET}= V \ostar A.
\end{aligned}
\end{equation}

The obtained heatmaps $\mathcal{H}_{CIET}$ contain the static and dynamic contextual information.  Hence, the total loss between  predicted and ground-truth heatmaps is the  $L_2$ loss of each joint which is given as,
\begin{equation}
\label{eqn:82}
\begin{aligned}
\Lb_{\mathcal{H}}=\sum\limits_{j \in N}||\mathcal{H}_{j_{CIET}}-\mathcal{H}_{j}^{*}||^2_{2}.
\end{aligned}
\end{equation}

\subsection{Hand Pose Estimation}\label{sec:3}
The output from CIET is fed to transformer encoder-decoder model. This model is different from the conventional transformer. We follow~\cite{carion2020end}, to predict the joints along with the hand identities. The joints with the hand identities are defined as $(h_i,j_i)$, where $h_i$ is the hand identity i.e. left or right hand, and $j_i$ is the joint location. The output from the transformer model is passed through a two-layer Multi-Layer Perceptron (MLP), followed by a linear projection layer and a softmax layer.  Distance between the predicted joints and the ground truth values is calculated using a cross-entropy loss function defined as
\begin{equation}
\label{eqn:84}
\begin{aligned}
\Lb_{joints}=\sum\limits_{i }CE((h_i,j_i),(h_i^*,j_i^*)),
\end{aligned}
\end{equation}

where  $(h_i,j_i)$ are the predicted values,   $(h_i^*,j_i^*)$ are the ground truth values and CE is the cross-entropy loss. Following~\cite{hampali2022keypoint},  we empirically set the threshold $\gamma=3$. If the re-projected joint is within  3 pixels, the joint is considered as the hand joint or else, ignored.

\begin{figure*}[t]\centering
\centering
\begingroup
\setlength{\tabcolsep}{0pt} 
\renewcommand{\arraystretch}{1.5}
\begin{tabular}{cccc}

\includegraphics[scale=.45]{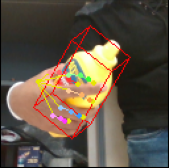}&
\includegraphics[scale=.45]{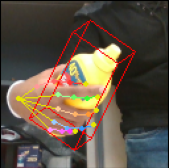}&
 
\includegraphics[scale=.45]{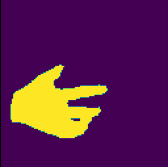}&
\includegraphics[scale=.45]{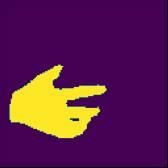}\\
(a) Predicted&(b) Ground Truth& (c) Predicted & (d) Ground Truth

\end{tabular}
\endgroup
\caption[ ]{Qualitative results for our method on the HO-3D. Here, (a), (b) shows Joint projections, and (c), (d) represents Hand Segmentation. }
\label{fig22}
\end{figure*}

The encoder and decoder consist of two attention sub-modules: sigmoid attention and softmax attention. The softmax attention sub-module is the same as in the conventional transformers with the softmax correlation map as ($C_{soft}$. The softmax attention $C_{soft}$ relates the information of each pixel of query, q with the key, $k$. The input to the encoder is the features $F$ obtained from the backbone network which are given as input to the Query (q), Key (k), and the Value (V) as shown in Fig.~\ref{fig3}. The correlation map ($C_{sig}$) is defined as 
\begin{equation}
\label{eqn:28}
\begin{aligned}
C_{sig}=sigmoid(\frac {qk^T}{\sqrt{d_k}}),
\end{aligned}
\end{equation}
where $d_k$ denotes the dimensions of the keys.  

The sigmoid attention filters the undesired high correlation by generating a correlation map between each query pixel and the global key information. Thus, we obtain the  final attention map ($C$) as 
\begin{equation}
\label{eqn:38}
\begin{aligned}
C=C_{soft}*C_{sig},
\end{aligned}
\end{equation}

The correlation map $C$ is used to extract the hand information from the occluded regions. After obtaining a combination of attention maps from both the attention modules, the multiplication of $C$ and the values $V$ takes place to get the final attention which is as follows:
\begin{equation}
\label{eqn:80}
\begin{aligned}
Attention=CV,
\end{aligned}
\end{equation}

Both these self-attentions are used for the Transformer encoder and decoder with an additional cross-attention (CA) module in the decoder of the transformer architecture that finally estimates the hand poses through a feed-forward network (FFN).

\subsection{Object Pose Estimation}
The object pose is estimated from the backbone network. These poses are then refined by passing through the CIET network similar to the joint heatmaps. A vector of 256 dimensions is obtained for each object point exactly similar to the hand pose. Similar to hand identity, object identity is also predicted which associates only the object key points from all the points originating from the object. The loss optimized for predicting the object pose is the symmetry-aware object corner loss which includes the corner of the 3D bounding box of the object in the rest pose as in~\cite{hampali2022keypoint}.

\subsection{Loss Optimization}
We have trained our model in an end-to-end fashion. The total loss for optimization is 
\begin{equation}
\label{eqn:8}
\begin{aligned}
\Lb=\Lb_{\mathcal{H}} + \Lb_{\mathcal{T}}+ \Lb_{joints}+ \Lb_{hand-pose}+ \Lb_{object-pose},
\end{aligned}
\end{equation}
where $\Lb_{\mathcal{T}}$, $\Lb_{hand-pose}$  and $\Lb_{object-pose}$are the $L1$ loss for relative translation between the two hands, the hand-pose and the object-pose loss as detailed in~\cite{hampali2022keypoint}.

\begin{table}[t]
\caption{Comparison on InterHand2.6M.}
\label{tab:15}  
\centering
\begin{tabular}{|m{7em}|m{3em}|m{3em}|m{3em}|m{4em}|}
\hline
\multirow{2}{*}{Methods}&\multicolumn{3}{c|}{MPJPE~(mm)}&MRRPE\\ \cline {2-4}
& Single& Two& All&(mm) \\ \hline
CNN+SA &13.53& 16.87& 15.31 &33.84\\
DETR~\cite{carion2020end}&12.81& 15.94& 14.48 &32.87\\
InterNet\cite{moon2020interhand2}&12.16& 16.02 &14.22 &32.57\\
Kypt Trans.~\cite{hampali2022keypoint}	& 11.25 & 14.73 & 14.21 & 31.87\\
\textbf{Ours}	& \textbf{9.82}&1\textbf{2.89}&\textbf{11.43}&\textbf{27.93}\\
\hline	
\end{tabular} 
\end{table}

\begin{table*}[t]
\caption{Comparison  on HO-3D V3 dataset. Note that ArtiBoost~\cite{li2021artiboost} uses additional training data which is not used in our method.}
\label{tab:1}  
\centering
\begin{tabular}{|c|c|M{9.5em}|M{10.1em}|M{10.2em}|M{8.6em}|}
\hline
Method  &Pose   Repr. & Joint Error (scale and trans. align.) in cms &Joint Error AUC (scale and trans. align.)& Joint Error (Procrustes align.) in cms & Joint Error AUC (Procrustes align.) \\ \hline
\cite{li2021artiboost}&2.5D &2.34& \textbf{0.565}& \textbf{1.08}& 0.785 \\
\cite{hampali2022keypoint}& 3D &2.98& 0.755& 1.43& 0.834\\
Ours&3D& \textbf{1.14 }&	0.723 &	1.20 & \textbf{0.518}\\
\hline
\end{tabular} 
\end{table*}

\begin{table*}[ht]
\caption{Ablation on HO-3D V3 dataset. }
\label{tab:3}  
\centering
\begin{tabular}{|c|M{3cm}|M{3.3cm}|M{3cm}|M{3cm}|}
\hline
Method  & Joint Error (scale and trans. align.) cms&Joint Error AUC (scale and trans. align.)& Joint Error (Procrustes align.) cms & Joint Error AUC (Procrustes align.) \\ \hline
w/ CIET &1.19 &	0.765 &	1.96&	0.500	\\
w/ sigmoid att.& 1.18 &	0.765 &	1.86 &	0.505 \\
w/ CIET \& sigmoid att.& 1.14 &	0.773 &1.20 &	0.518  \\
\hline	
\end{tabular} 
\end{table*}

\begin{table}[ht]
\caption{Comparison  on H$_2$O3D.}
\label{tab:12}  
\centering
\begin{tabular}{|M{6.7em}|M{5.7em}|M{5.7em}|M{4.9em}|}
\hline
Method&MPJPE~(cm)&MRRPE~(cm)&MSSD~(cm)\\ \hline
Kypt Trans.~\cite{hampali2022keypoint}	& 3.42 & 8.55 & 8.37\\
Ours& \textbf{2.86 }&\textbf{7.22}&\textbf{7.05}\\
\hline	
\end{tabular} 
\end{table}

\section{Results and Discussions}
The model is implemented in the PyTorch framework on NVIDIA 1080 GPU. We have evaluated our model on three hand pose datasets: InterHand2.6M~\cite{moon2020interhand2}, HO-3D\_V3~\cite{hampali2020honnotate}  and  H$_2$O-3D~\cite{hampali2022keypoint}. The model is trained for  30 epochs on InterHand2.6M and 150 epochs on    HO-3D\_V3 and H$_2$O-3D dataset. Results for Kypt Trans.~\cite{hampali2022keypoint} are reproduced keeping the setting use\_big\_decoder~=~False due to GPU constraints. 

\subsection{Comparative Analysis}
We compare the results of our method on the InterHand2.6M dataset with the state-of-the-art methods, and the baseline (CNN + SA), when using the 2.5D pose representation. In this paper, all the experiments and evaluations are performed for a dataset obtained at 5 fps which contains 1.36M train images and 849K test images with each image having $512\times334$ resolution. The CNN + SA model first extracts the low-level CNN features which are given as input to the Transformer self-attention (SA) module. The output pose is predicted using an MLP model. The comparative results are shown in Table~\ref{tab:15}. Our method achieves 11\% higher accuracy than Keypoint Transformer~\cite{hampali2022keypoint} and about 23\%  rise in accuracy is seen from the InterNet~\cite{moon2020interhand2} model. This shows that the contextual information extracted using CIET helps to extract and estimate more accurate hand pose.

We have evaluated the proposed method for the Ho3D dataset on the 3D joint representation and the comparative results are shown in Table~\ref{tab:1}. We compare our results with ArtiBoost~\cite{li2021artiboost} and the Keypoint transformer~\cite{hampali2022keypoint}. ArtiBoost~\cite{li2021artiboost} uses 2.5D for pose representation while Keypoint transformer~\cite{hampali2022keypoint} uses 3D pose. Following~\cite{hampali2022keypoint}, we have also used 3D for pose representation.  From Table~\ref{tab:1}, it is seen that our method outperforms other methods when scale and translation alignment are considered. We also achieve a MSSD value of 9.4535 cm. We show qualitative results in Fig.~\ref{fig22}.

Table~\ref{tab:12} shows the comparative results obtained on the H$_2$O-3D dataset. Our method achieves MPJPE of 2.86cm and MRRPE of 7.22cm. Since this dataset contained interacting hands, it faces large occlusion, and estimating 3D hand pose becomes more challenging in this dataset. So, comparing Table~\ref{tab:15} and Table~\ref{tab:12}, we can conclude that it is easy to estimate 3D hand pose for InterHand2.6M while for H$_2$O-3D dataset, this task is very challenging due to interacting hands and occlusion faces from objects. From  Table~\ref{tab:12}, when comparing our method with~\cite{hampali2022keypoint}, it is seen that our method achieves about 8\% higher joint accuracy and about 14\% higher accuracy when translation is involved. Similarly, for object pose estimation our method outperforms~\cite{hampali2022keypoint} by obtaining   MSSD values of 7.05 cm.

\subsection{Ablation Study}
We perform an ablation study of our model on the HO-3D\_V3 dataset.  Since our model uses a CIET network that extracts the contextual information from the input, we perform ablation for the CIET network, sigmoid activation attention, and softmax attention. Therefore,  a total of three possible networks are used for studying the model architecture. First model includes only CIET network, without using the sigmoid attention, second considers only the sigmoid attention without the CIET network and last model includes both CIET and the sigmoid attention module in the decoder of the transformer for predicting the hand-object pose.   

The ablation results are presented in Table~\ref{tab:3} and from the table it is very clear that the best results are obtained when both CIET and sigmoid attention modules are used. However, joint error accuracy after Procrustes alignment performs best when only CIET is employed without the sigmoid attention.  

\section{Conclusion}
We have presented a novel method OccRobNet for interacting hand pose estimation.  Our network is robust against occlusion as it estimates hand pose even when the hand is occluded with the object or a complex two-hand interaction is performed. Our model utilizes contextual information to enhance the extracted hand information from the CNN backbone network. This contextual information could be extracted even when the hand faces, which makes the complete system robust against occlusion. Secondly, sigmoid attention is used which filters out the undesirable correlation between the pixels and helps the complete system to obtain the state-of-the-art results on three challenging hand pose datasets.

\bibliography{A} 
\bibliographystyle{ieeetr}

\end{document}